\documentclass[sigconf]{acmart}
\usepackage{multirow} % new add
\usepackage{float}
\usepackage{subfig}                 %
\usepackage{makecell}
\AtBeginDocument{%
  }

\copyrightyear{2024}
\acmYear{2024}
\setcopyright{acmlicensed}\acmConference[CIKM '24]{Proceedings of the 33rd ACM International Conference on Information and Knowledge Management}{October 21--25, 2024}{Boise, ID, USA}
\acmBooktitle{Proceedings of the 33rd ACM International Conference on Information and Knowledge Management (CIKM '24), October 21--25, 2024, Boise, ID, USA}
\acmDOI{10.1145/3627673.3679618}
\acmISBN{979-8-4007-0436-9/24/10}

\begin{document}

\title{SVIPTR: Fast and Efficient Scene Text Recognition with Vision Permutable Extractor}

\author{Xianfu Cheng}
\orcid{0000-0003-1130-8302}
%\email{webmaster@marysville-ohio.com}
\affiliation{%
  \institution{CCSE, Beihang University}
  \city{Beijing}
  \country{China}
  % \state{Beijing Shi}
}
\email{buaacxf@buaa.edu.cn}

\author{Weixiao Zhou}
\orcid{0009-0006-8929-0834}
\affiliation{%
  \institution{CCSE, Beihang University}
  \city{Beijing}
  \country{China}
  }
\email{wxzhou@buaa.edu.cn}

\author{Xiang Li}
\orcid{0009-0004-2003-9217}
\affiliation{%
  \institution{Beihang University}
  \city{Beijing}
  \country{China}
  }
\email{xlggg@buaa.edu.cn}

\author{Jian Yang}
\orcid{0000-0003-1983-012X}
\affiliation{%
  \institution{Beihang University}
  \city{Beijing}
  \country{China}
  }
\email{jiaya@buaa.edu.cn}

\author{Hang Zhang}
\orcid{0009-0009-3310-079X}
\affiliation{%
  \institution{CCSE, Beihang University}
  \city{Beijing}
  \country{China}
  }
\email{zhbuaa0@buaa.edu.cn}

\author{Tao Sun}
\orcid{0009-0006-8694-8530}
\affiliation{%
  \institution{CCSE, Beihang University}
  \city{Beijing}
  \country{China}
  }
\email{ASC_8384@foxmail.com}

\author{Wei Zhang}
\orcid{0009-0001-4419-4551}
\affiliation{%
  \institution{CCSE, Beihang University}
  \city{Beijing}
  \country{China}
  }
\email{zwpride@buaa.edu.cn}

\author{Yuying Mai}
\orcid{0009-0003-3448-1746}
\affiliation{%
  \institution{Beijing Jiaotong University}
  \city{Beijing}
  \country{China}
  }
\email{maiyy1221@163.com}

\author{Tongliang Li}
\orcid{0000-0002-2488-2787}
\authornote{Corresponding Author.}
\affiliation{%
  \institution{Beijing Information Science and Technology University}
  \city{Beijing}
  \country{China}
  }
\email{tonyliangli@bistu.edu.cn}

\author{Xiaoming Chen}
\orcid{0009-0000-9314-3753}
\affiliation{%
  \institution{Shenzhen Intelligent Strong Technology Co.,Ltd.}
  \city{Shenzhen}
  \country{China}
  }
\email{chenxiaoming@aistrong.com}

\author{Zhoujun Li}
\orcid{0000-0002-9603-9713}
%\authornote{Corresponding Author.}
\authornotemark[1]
\affiliation{%
  \institution{CCSE, Beihang University}
  \city{Beijing}
  \country{China}
  }
\email{lizj@buaa.edu.cn}
\renewcommand{\shortauthors}{Xianfu Cheng et al.}

\begin{abstract}
  Scene Text Recognition (STR) is an important and challenging upstream task for building structured information databases, which involves recognizing text within images of natural scenes. Although current state-of-the-art (SOTA) models for STR exhibit high performance, they typically suffer from low inference efficiency due to their reliance on hybrid architectures comprised of visual encoders and sequence decoders. In this work, we propose a \textbf{VI}sion \textbf{P}ermutable extractor for fast and efficient \textbf{S}cene \textbf{T}ext \textbf{R}ecognition (SVIPTR), which achieves an impressive balance between high performance and rapid inference speeds in the domain of STR. Specifically, SVIPTR leverages a visual-semantic extractor with a pyramid structure, characterized by the Permutation and combination of local and global self-attention layers. This design results in a lightweight and efficient model and its inference is insensitive to input length. Extensive experimental results on various standard datasets for both Chinese and English scene text recognition validate the superiority of SVIPTR. Notably, the SVIPTR-T (Tiny) variant delivers highly competitive accuracy on par with other lightweight models and achieves SOTA inference speeds. Meanwhile, the SVIPTR-L (Large) attains SOTA accuracy in single-encoder-type models, while maintaining a low parameter count and favorable inference speed. Our proposed method provides a compelling solution for the STR challenge, which greatly benefits real-world applications requiring fast and efficient STR. The code is publicly available at https://github.com/cxfyxl/VIPTR.
\end{abstract}

\begin{CCSXML}
<ccs2012>
   <concept>
       <concept_id>10010147.10010178.10010224.10010245.10010251</concept_id>
       <concept_desc>Computing methodologies~Object recognition</concept_desc>
       <concept_significance>500</concept_significance>
       </concept>
   <concept>
       <concept_id>10010147.10010257.10010293.10010294</concept_id>
       <concept_desc>Computing methodologies~Neural networks</concept_desc>
       <concept_significance>500</concept_significance>
       </concept>
   <concept>
       <concept_id>10010147.10010178.10010224.10010225.10010227</concept_id>
       <concept_desc>Computing methodologies~Scene understanding</concept_desc>
       <concept_significance>500</concept_significance>
       </concept>
   <concept>
       <concept_id>10010147.10010178.10010224.10010240.10010241</concept_id>
       <concept_desc>Computing methodologies~Image representations</concept_desc>
       <concept_significance>500</concept_significance>
       </concept>
 </ccs2012>
\end{CCSXML}

\ccsdesc[500]{Computing methodologies~Object recognition}
\ccsdesc[500]{Computing methodologies~Neural networks}
\ccsdesc[500]{Computing methodologies~Scene understanding}
\ccsdesc[500]{Computing methodologies~Image representations}
\keywords{Scene Text Recognition, Attention mechanism, Vision Transformer, Visual-semantic analysis, Length-Insensitive}
\iffalse
\begin{teaserfigure}
  \includegraphics[width=\textwidth]{sampleteaser}
  \caption{Seattle Mariners at Spring Training, 2010.}
  \Description{Enjoying the baseball game from the third-base
  seats. Ichiro Suzuki preparing to bat.}
  \label{fig:teaser}
\end{teaserfigure}
\fi


\maketitle

\section{Introduction}
Text Recognition (TR) is crucial for multimodal document analysis and building structured databases, autonomous driving, office automation, and other applications~\cite{shi2016end,lvpm3,alm,m2c,str1,str2,mp4}. Recent deep learning advancements have spurred growth in Scene Text Recognition (STR), which aims to digitize text from images in natural settings for use in language processing tasks. Despite progress, STR still faces hurdles such as distorted text, varied fonts, shadows, and complex backgrounds. Efforts are ongoing to improve recognition precision, with an emphasis on creating models that are accurate, fast, and lightweight for versatile deployment.

\begin{figure}[!t]
	\centering
        %\vspace{10pt}
	\includegraphics[width=1.0\linewidth]{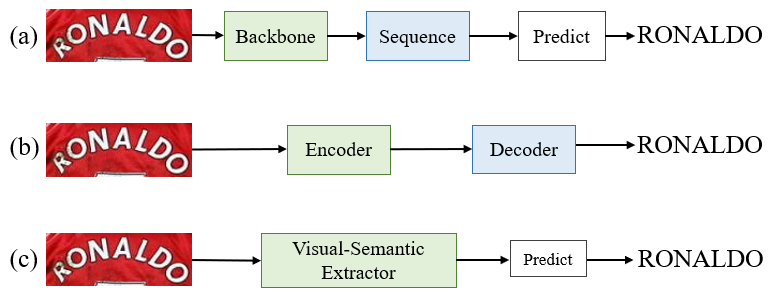}
        %\vspace{0pt}
	\caption{Model Architecture Evolution for Scene Text Recognition.}
	\label{fig:STRarch}
        %\vspace{10pt}
        \Description{Model Architecture Evolution for Scene Text Recognition.}
\end{figure}

STR initially relied on deep learning models like CRNNs~\cite{shi2016end} that combined visual feature extraction with sequence processing (as shown in Figure \ref{fig:STRarch}(a)), favored for their fast inference despite sensitivity to text disturbances. Subsequent models adopted encoder-decoder, autoregressive approaches (as shown in Figure \ref{fig:STRarch}(b)), enhancing accuracy through contextual understanding but at the cost of slower word-by-word translation speeds. More recent advances involve visual semantic parsing frameworks (as shown in Figure \ref{fig:STRarch}(c)) that integrate semantic knowledge to improve recognition accuracy in specific contexts, which often require large models or intricate semantic systems, leading to efficiency compromise.

Recent researchers~\cite{hu2020gtc,zhang2023self} tried to develop simple and effective model structures by using the attention mechanism to aggregate the sequence features corresponding to the same character, or regularising the extracted sequence frame by frame for supervised learning of individual characters. These strategies are used to align the Cross-Entropy (CE) loss and Connectionist Temporal Classification (CTC) loss, on the one hand, the convergence speed of CTC loss is improved without affecting inference speed, and on the other hand, the character and sequence features are strengthened. From the perspective of simplifying the model architecture, some models used only CNN~\cite{borisyuk2018rosetta} or ViT~\cite{atienza2021vision} for identification, which is efficient in inference but has a large gap in accuracy compared to SOTA methods. Recent researchers~\cite{wang2022multi,cheng2023lister,fujitake2024dtrocr} have proposed a series of innovative optimization methods for feature sequence decoders, simplifying the encoder-decoder architecture and effectively promoting a balance between high performance and inference efficiency. In addition, inspired by the pyramid structure of Swin Transformer~\cite{liu2021swin}, the SVTR model~\cite{du2022svtr}, designed for single visual modality input, is proposed to simultaneously process the implicit local visual information in text images and unsupervised mining of global contextual semantic information.

The STR models based on vision Transformer set a typical example for fast and effective STR schemes for a single visual modality. On one hand, it practices the effectiveness of replacing convolution with a self-attention mechanism to realize a new visual feature extractor. On the other hand, it reveals the importance of modeling global dependencies of lateral symbol sequences in text image recognition tasks, which even replace bidirectional LSTMs~\cite{sun2022neural} and sequence decoders. Despite the outstanding performance of current ViT-based STR, it is clear that it still has several shortcomings: (1) The computation of self-attention relies on absolute position encoding and fixed-size mask operation, which limits the length of the input image; (2) In each stage, only one type of global and local relations is mined separately, and the distance between the steps of global dependencies is limited to save reasoning time, so the recognition effect of long text images is not good; (3) The Resource occupancy during training and CPU inference time consumption of models in general scenarios are still larger than that of CRNN with the same parameter magnitude.

To further enhance the vision model and meticulously control computational complexity to achieve more efficient STR while maintaining high accuracy, we propose SVIPTR—a visual-semantic-based approach for a faster, more accurate, and cross-language universal STR model. Inspired by Cross-Shaped Windows Vision Transformer (CSwin)~\cite{dong2022cswin} and other visual models based on enhanced self-attention mechanisms, SVIPTR first divides the text image evenly into multiple 2D patches, each patch contains part of the characters or background. This way of dividing patches facilitates the simultaneous extraction of local features of characters and the capture of potential contextual semantic dependencies between different characters. In each token mixing attention module, we try to use conditional positional embedding (CPE)~\cite{chu2021conditional} combined with Manhattan self-attention (MaSA)~\cite{fan2023rmt} and CSWin Attention to explore the local information of characters, and use Overlapping Spatial Reduction Attention (OSRA)~\cite{lou2023transxnet} module to model global dependencies. Specifically, along the horizontal direction Model the global dependency between characters in the axial direction, model the global relationship from top to bottom of the same character along the vertical axis, and align to achieve weight sharing. Compared with SWin, this method significantly reduces the time complexity and does not require the use of a fixed-size mask group to limit the length of the input image. At the same time, we also use a pyramid structure for multi-stage local and global information hybrid modeling, which greatly improves the feature extraction capabilities of text images. To sum up, SVIPTR can ensure that the recognition effect and inference efficiency exceed the advanced vision Transformer-based recognition model at a smaller model scale. 

The main contributions of this work can be summarized as:
\begin{itemize}
        \item We verify that the use of sparse operators and different Permutations and combinations of self-attention mechanisms can accelerate the calculation of Vision Transformer, and the model built based on these mechanisms can still achieve comparable accuracy to high-level vision-language models in STR tasks. Integrated, achieving the balance between performance and speed advantages that STR has been pursuing.
	
	\item We proposed SVIPTR, a visual-semantic feature extraction model tailored for parsing image text. The model can accurately recognize cross-language image text input and is insensitive to the length of image input, which has a good application prospect.
	
	\item We verified the superiority of SVIPTR on cross-language benchmark datasets and manually annotated industrial application datasets. Among them, SVIPTRv1-L achieved higher accuracy than other ViT-based encoder models on both Chinese and English STR. SVIPTRv1-L combined with BiLSTM decoder achieves the SOTA level on STR benchmarks. SVIPTRv2-T achieves the most efficient inference while ensuring accuracy. It is inferred on an NVIDIA V100 GPU with a parameter of 5.1M and consumes an average of 3.3ms per text image.

\end{itemize}

\section{Related Work}
\begin{figure*}[!ht]
	\centering
	\includegraphics[width=0.95\textwidth]{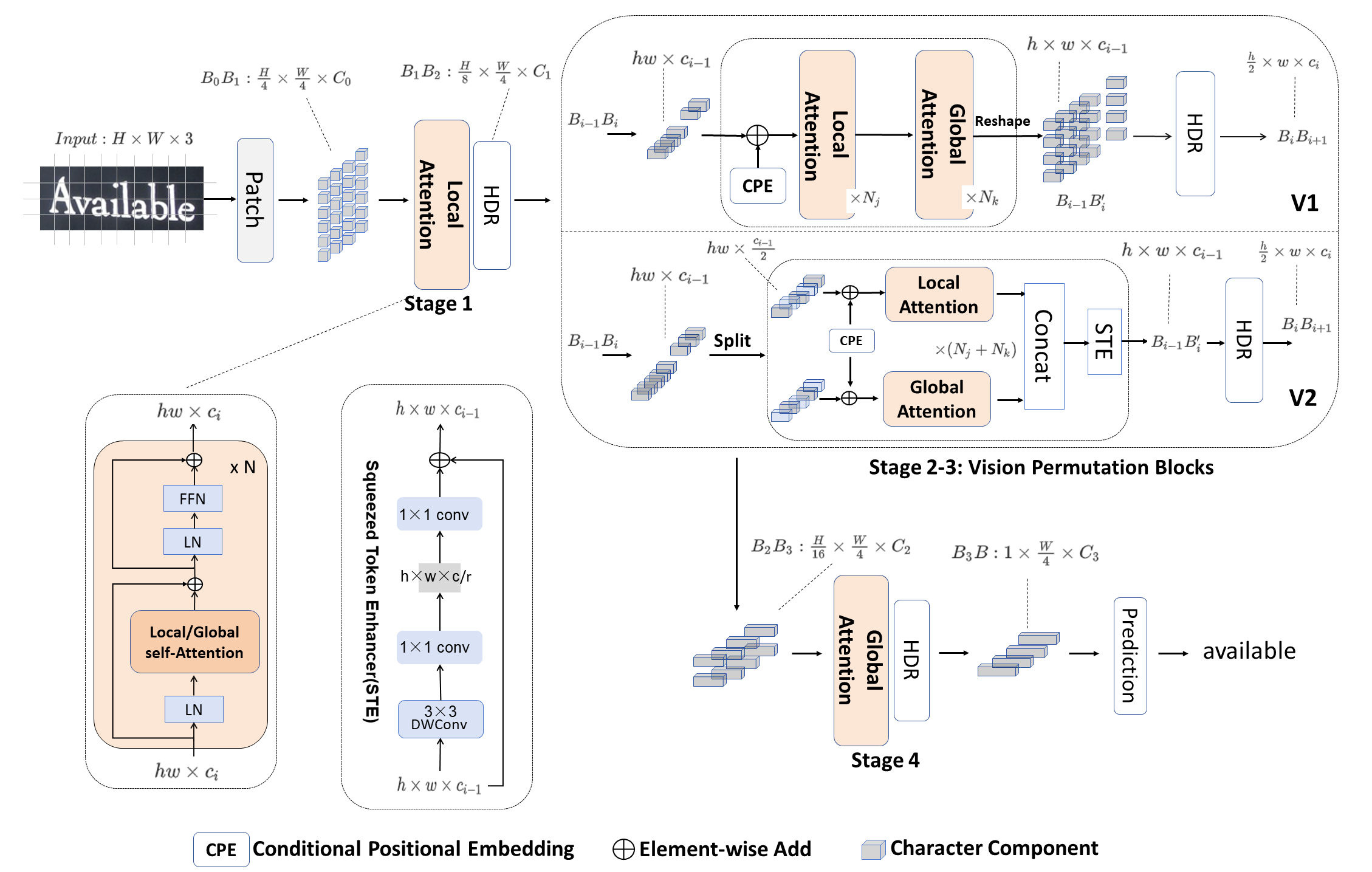}
        %\vspace{0pt}
	%\hfill
	\caption{Overall architecture of the proposed SVIPTR. It is a four-stage network with three-stage height progressively decreasing. In each stage, a series of attention-mixing blocks are carried out and followed by a subsampling or combining operation. In particular, SVIPTR designs two visual feature fusion modes in vision permutation blocks: series and parallel. At last, the recognition is conducted by the CTC decoder.}
        \Description{Overall architecture of the proposed SVIPTR. It is a four-stage network with three-stage height progressively decreasing. In each stage, a series of attention-mixing blocks are carried out and followed by a subsampling or combining operation. In particular, SVIPTR designs two visual feature fusion modes in vision permutation blocks: series and parallel. At last, the recognition is conducted by the CTC decoder.}
	\label{fig:SVIPTRarch}
\end{figure*}

\subsection{CNN-Based Vision Modules.}
Convolutional neural networks (CNN) serve as pivotal tools for extracting visual information in previous tasks. VGG~\cite{simonyan2014very} demonstrated that a series of 3x3 convolutions is sufficient to train a high-performance model. Later, ResNet~\cite{he2016deep} introduced residual connections along with the batch normalization (BN) layer, which enabled the training of exceedingly deep neural networks and further improved performance. In the realm of STR, the CNN-based vision model transforms a pyramid structure capable of processing local features across multiple stages. This transformation effectively compresses the two-dimensional image input into a one-dimensional feature sequence, allowing for decoding the sequence without reliance on recurrent neural networks (RNN). High recognition accuracy can also be obtained, which proves the effectiveness of locality bias extracted by CNN. We believe that the experience gained in CNN-based vision models, including the pyramid structure for extracting locality bias through small kernel convolutions, residual connections, and step-by-step downsampling, remains valid in current research. Therefore, these techniques continue to be utilized in our proposed STR architecture.

\subsection{Transformer-Based Vision Modules.}
ViT~\cite{dosovitskiy2020image} is the first work to introduce Transformer~\cite{vaswani2017attention,crop,touvron2021training} into the construction backbone of vision tasks. Experiments demonstrate that, with sufficient training data, ViT outperforms traditional CNNs in vision tasks. DeiT~\cite{touvron2021training} introduces a new training strategy, enabling ViT to achieve superior performance compared to CNNs on small-scale training sets such as ImageNet-1K. These methods demonstrate the efficacy of global dependency modeling through self-attention mechanisms. Later, the pyramid structure also enhances the performance of Transformer-based models across various vision tasks. Based on the pyramid structure, Swin Transformer proposes to use self-attention within each local window. This can be regarded as a use of locality bias. Vision Transformers are integrated into STR tasks as visual feature encoders, leading to a significant improvement in accuracy. VisionLAN~\cite{wang2021two}, SVTR, MGP-STR~\cite{wang2022multi} and MaskOCR~\cite{lyu2022maskocr} have all come close to the SOTA effect. 
% Although the self-attention mechanism has greatly improved the effect of text recognition, its computational complexity is quadratic compared to the number of input tokens. As the resolution of the input image increases, the inference time consumption also increases quadratically. 

\subsection{Sparse Self-attention and Spatial Modeling.}
A challenging problem in the design of Transformer is that the computational cost of global self-attention is very expensive~\cite{vaswani2017attention,xmt,ganlm}, while local self-attention usually limits the interaction field of each token, resulting in insufficient information mining. The CSWin Self-Attention mechanism uses parallel computing to form the self-attention of the horizontal and vertical stripes of the cross-shaped window, achieving powerful feature modeling capabilities while limiting computing costs. Decomposed MaSA achieves efficient and significant feature representation by defining a spatial attention attenuation matrix and its decomposition paradigm based on two-dimensional Manhattan distance, effectively layering local information on a global scale. Spatial reduction attention (SRA)~\cite{wang2021pyramid} effectively extracts global information by utilizing sparse token-region dependencies, and OSRA introduces overlapping spatial reduction (OSR)~\cite{lou2023transxnet} into SRA to achieve better performance by using larger and overlapping patches. 

\section{Method}
\subsection{Design Guidelines}
In this study, we use some important properties of CNNs, ResNet, and several self-attention mechanisms in a new architecture inspired by hierarchical vision Transformers to design a paradigm of high-performance networks for STR. Here are the design guidelines we followed:
\begin{enumerate}
\item Explicitly vectorizing the character features in the form of segmented chunks and injecting them into the front of the network.
\item Design permutation and combination modes of self-attention mechanism to effectively fuse visual and semantic information of characters embeddings in the middle layer.
\item Implicitly handling the length information of input images during network initialization and designing shape-insensitive self-attention mechanisms to replace all modules that depend on the input size.
\end{enumerate}

\subsection{Overall Architecture}
The schematic representation of the proposed SVIPTR is depicted in Figure \ref{fig:SVIPTRarch}. It is a four-stage network with three height progressively decreased layers. Given an image text of size $H\times W\times 3$, it undergoes an initial transformation into $\frac{H}{4}\times \frac{W}{4} \times C_0$ patches through a two-layer CNN-based patch embedding. Since each patch is associated with a text character or part of the background in the image, these patches are called character components. Subsequently, four stages of feature extraction are performed at different scales, each stage consists of a series of self-attention-based hybrid blocks, followed by dimension reduction in half using convolutional or global average pooling operations along the height dimension. Finally, the extractor generated a representation called $B$ with a size of $1\times \frac{W}{4}\times C_3$, which captures multi-granularity characters' sequence features. At the same time, the CTC decoder is used to linearly predict $B$ to obtain the recognition result.

% Finally, the CTC decoder is used to perform linear prediction and deduplication on the character sequence to obtain the recognition result.

% Both Local and Global Attention blocks are employed to concentrate on character feature extraction and capture dependencies between character components.

% serves to characterize character component features at different distances and multi-scales and the contextual dependencies between character components, and

\subsection{Patch Embedding}
The first step of the vision Transformer is to cut the image into several equal-sized patches, and then hand them over to the subsequent self-attention module for processing in the form of several sets of sequences. In this article, for an image text, first convert it from $X\in R^{H\times W\times 3}$ into several character components $B_0B_1\in R^{\frac{H}{4}\times \frac{W}{4}\times C_0}$. Alternatively, we implement the patch embedding by using two consecutive $3\times 3$ convolutions with stride $2$ and BNs. This scheme~\cite{du2022svtr} gradually increases feature dimensions, which is beneficial to feature fusion.

\subsection{Character and String Modeling}
%End-to-end text recognizers rely heavily on the mining of character features and string context. However, 
Existing studies mainly use feature sequences with height $1$ to represent image text. Each feature corresponds to a small local region. However, these segments are usually noisy, especially for irregular text, failing to describe the characters of an image well. In contrast, vision Transformer introduces two-dimensional feature representation, which not only enables more accurate discrimination between different characters and backgrounds but also fully exploits the potential contextual dependencies in patch sequence.
% In contrast, the Vision Transformer introduces a two-dimensional feature representation, which enables more accurate discrimination between different characters and backgrounds. This suggests that two-dimensional features can be used to effectively represent the local strokes of characters and the potential contextual dependencies of strings.

\begin{figure}[!t] % htbp
	\centering
	\subfloat[]{
		\includegraphics[width=0.45\columnwidth]{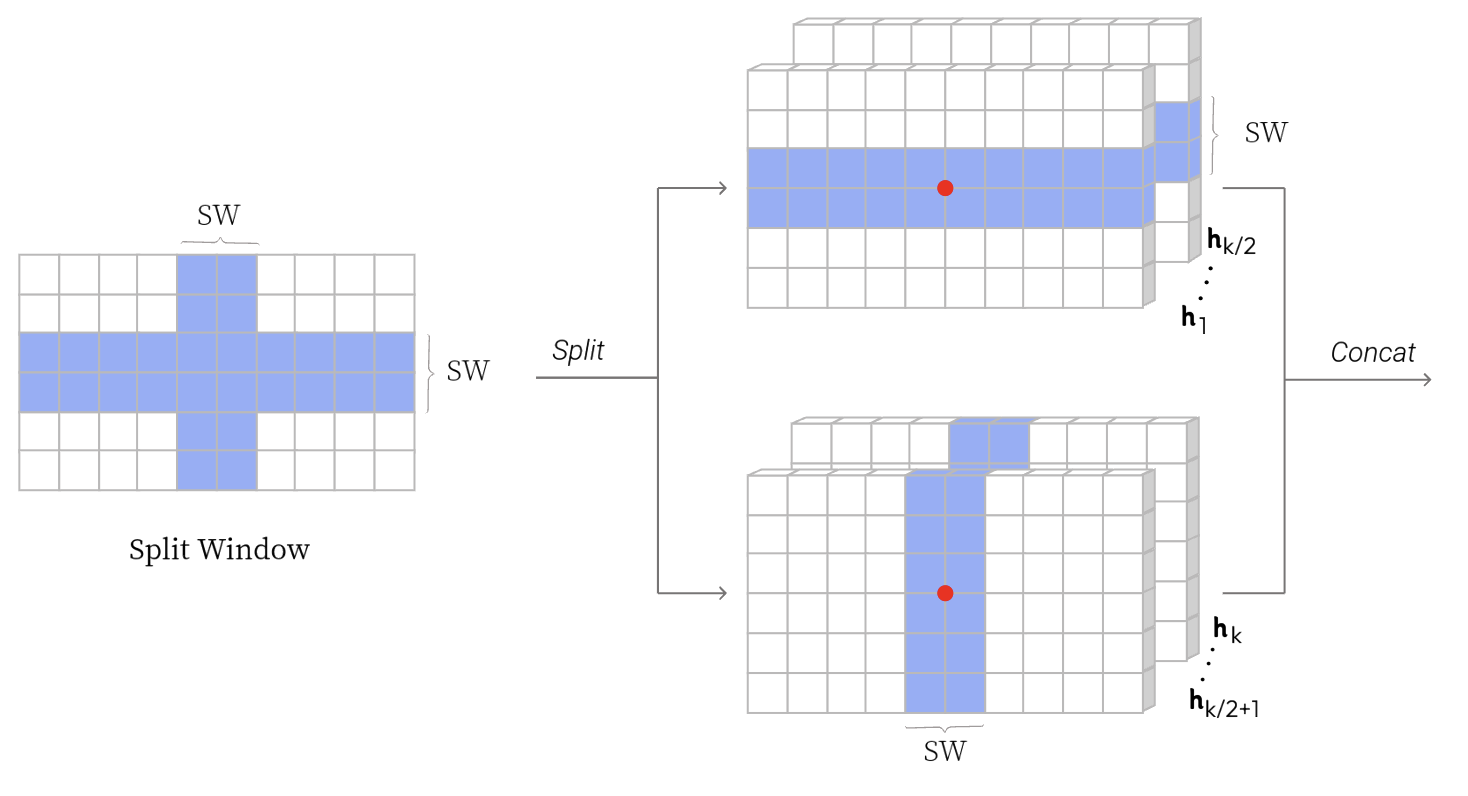}
		\label{subfig:cswin}
	}\hspace{5pt}
	\subfloat[]{
		\includegraphics[width=0.45\columnwidth]{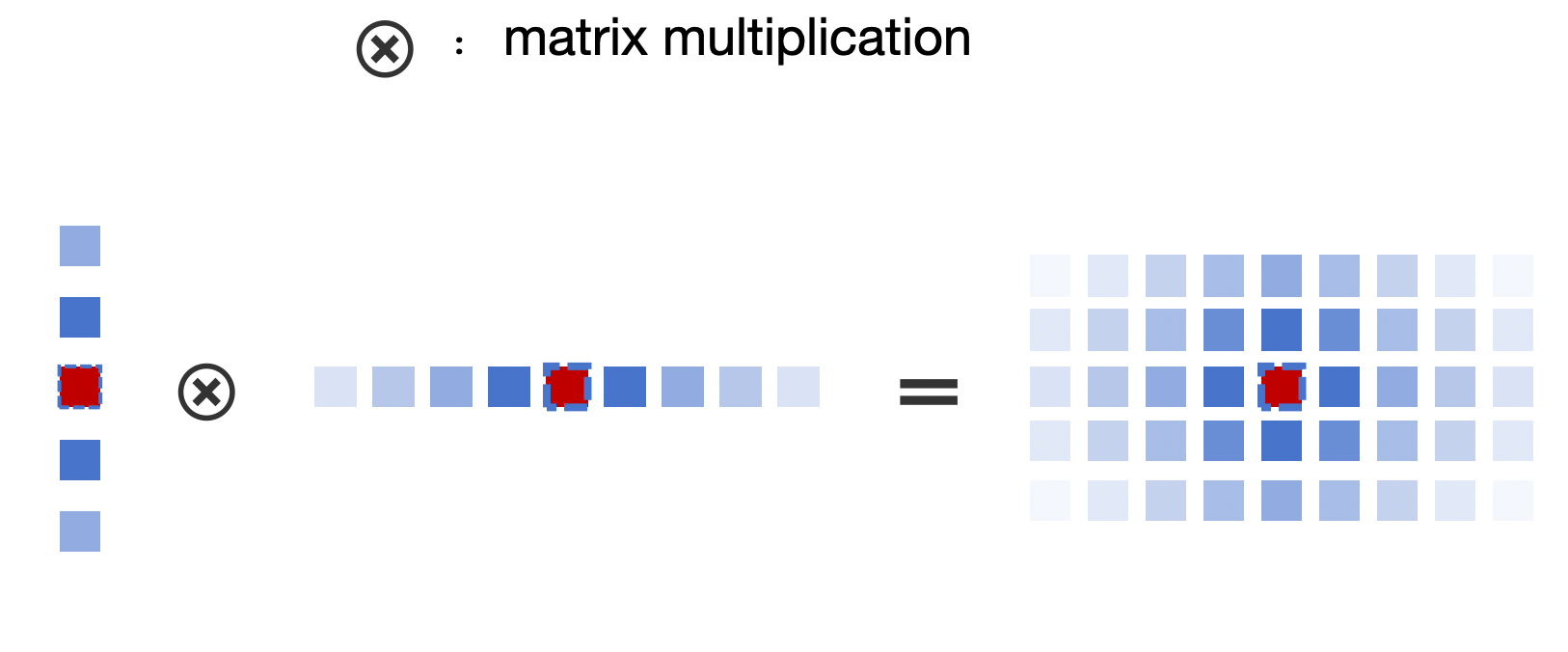}
		\label{subfig:dmsa}
	}\\
	\subfloat[]{
		\includegraphics[width=0.45\columnwidth]{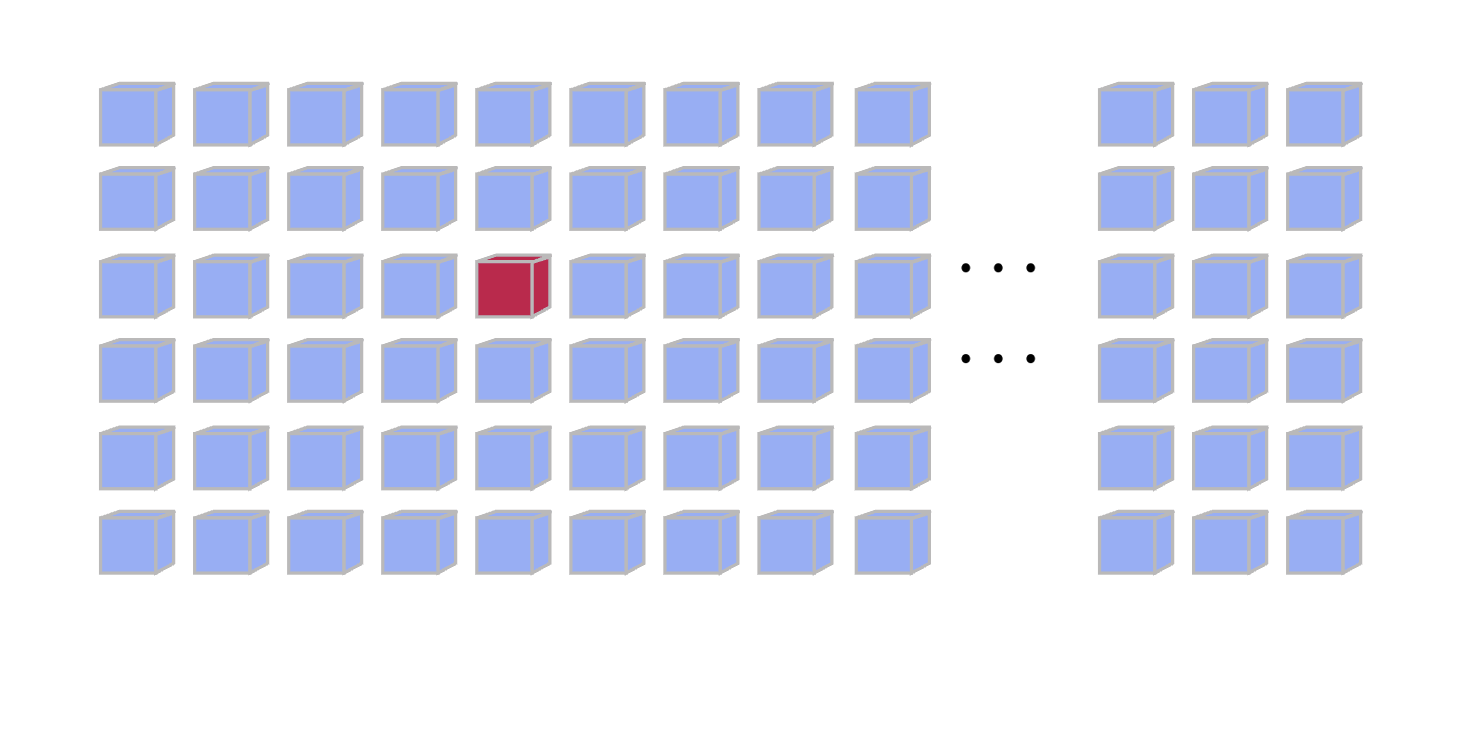}
		\label{subfig:mhsa}
	}\hspace{5pt}
	\subfloat[]{
		\includegraphics[width=0.45\columnwidth]{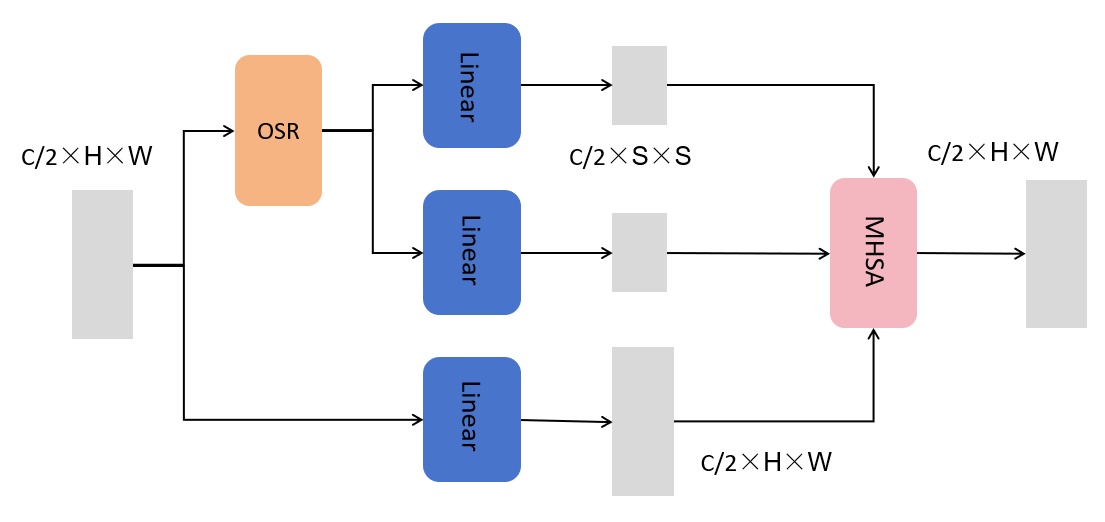}
		\label{subfig:osra}
	}
	\caption{Illustration of (a) Cross-Shaped Windows Self-Attention (CSWin), (b) Decomposed Manhattan Self-Attention (D-MaSA), (c) Multi-Head Self-Attention (MHSA), and (d) Overlapping Spatial Reduction Attention (OSRA).}
        \Description{Illustration of (a) Cross-Shaped Windows Self-Attention (CSWin), (b) Decomposed Manhattan Self-Attention (D-MaSA), (c) Multi-Head Self-Attention (MHSA), and (d) Overlapping Spatial Reduction Attention (OSRA).}
	\label{fig:multiSA}
        %\vspace{15pt}
\end{figure}

More specifically, image text requires complementary features to be extracted from two perspectives. 
The first involves local visual features, representing internal associations and morphological features of each character. The second is contextual semantic relations, such as correlations between different characters or between text and non-text components. Therefore, we design two classes of self-attention blocks with different receptive fields to sense relevance. Mathematically speaking, a sequence of character components $B_{i-1}B_i$ in the previous stage, is first reshaped as a sequence of features. Upon inputting the self-attention block, Layer Normalization (LN) and Position Embedding (PE) are first applied. Subsequently, the self-attention mechanism is employed for dependency modeling.
Finally, LN and Feed Forward Network (FFN) are successively applied for feature fusion, with the residual connection applied.

\textbf{Positional Embedding.} Since self-attention is permutation invariant and ignores tokens' positional information, positional Embedding (PE) is widely used in Transformers to add such information back. Typical positional Embedding mechanisms include absolute positional encoding (APE)~\cite{vaswani2017attention}, relative positional encoding (RPE)~\cite{liu2021swin}, and CPE. APE and RPE are designed for a specific input size and are not friendly to varying input resolutions. CPE takes the feature as input and can generate the positional encoding for arbitrary input resolutions. Then the generated positional encoding will be added to the input feature. Locally-enhanced Positional Encoding (LePE)~\cite{dong2022cswin} shares a similar spirit as CPE, but proposes to add the positional encoding as a parallel module to the self-attention operation and operates on projected values in each Transformer block.

\textbf{Local Attention.} The goal of Local Attention is to encode character morphological features and establish correlations within character components, thereby simulating stroke features. Local Attention needs to consider the neighborhood of character components on the two-dimensional coordinate system. Drawing on the design ideas of MaSA and CSWin Self-Attention, we designed two efficient self-attention mechanisms that explicitly model local features on a two-dimensional coordinate system. MaSA is shown in Figure \ref{fig:multiSA}\subref{subfig:dmsa}. The spatial attenuation matrix of the image means that character features are highlighted globally and background features are weakened. CSWin Self-Attention is shown in Figure \ref{fig:multiSA}\subref{subfig:cswin}. This mechanism is implemented by executing local self-attention of horizontal and vertical stripes of step size called Split-Window ($SW$) decomposed from a certain cross window in parallel. For image text input, considering that the image resolution is larger than the Channel in the early stage, we choose a smaller $SW$ in the early stage, and because the size of the input image text changes dynamically during processing, and the length and width resolution needs to be guaranteed. Divisible by $SW$, we empirically set $SW$ for stages using CSWin Self-Attention to $1$ or $2$ by default.

\textbf{Global Attention.} As shown in Figure \ref{fig:multiSA}\subref{subfig:mhsa}, Global Attention calculates the contextual dependencies between all character components. Since text and non-text are the two main elements in an image, this universal blend can create long-distance dependencies between the components of different characters. In addition, using the self-attention mechanism can also weaken the influence of non-text components while enhancing the importance of text components. In addition, to save computational costs, we introduced OSRA, as Figure \ref{fig:multiSA}\subref{subfig:osra}, which adds overlapping convolution operations to MHSA to sparse character components in some channels into more salient features without losing edge information. The use of salient features greatly reduces the number of tokens required to calculate the self-attention mechanism, thus improving computational efficiency and performance.

% As mentioned above, the hybrid blocks in SVIPTR are designed to extract distinct yet complementary features. At each stage, Local and Global Attention are recurrently applied multiple times for comprehensive feature extraction. Effect of the two kinds of blocks Permutations can be seen in ablation experiments.
% In this way, blocks are initially processed locally before being addressed globally.

\subsection{Height Dimension Reduction}
To save computational costs and extract significant features layer by layer, we added the Height Dimension Reduction operation (HDR) after each self-attention block in every stage of SVIPTR, like classic CNN backbones~\cite{borisyuk2018rosetta}. Using the output of the features from the previous Attention block, we first reshape it into an embedding of size $H\times W\times C_{i-1}$, representing the current height, width, and channel respectively. In the previous stage, we then employ a $3\times 3$ convolution with a height dimension stride $2$ and a width dimension stride $1$, followed by LN, producing an embedding of size $\frac{H}{2}\times W\times C_i$. In the final stage, we first use AdaptiveAvgPool to compress the height dimension to $1$, followed by Fully Connected Layers, Hardswish Activation, and Dropout. By doing so, the character components are further compressed into feature sequences, where each element is represented by a feature of length $C_3$. Using AdaptiveAvgPool avoids applying convolution to embeddings with very small dimensions, such as height $2$, while being able to apply a variety of initial input heights.

The HDR block halves the height while maintaining a constant width. It not only reduces the computational cost but also compresses the height dimension to establish a multi-scale representation for each character without affecting the token layout in the width dimension. We also increase the channel dimension $C_i$ to compensate for the loss of information.

\subsection{Prediction}
We finally use the CTC decoder to perform parallel linear prediction on the obtained combined features to achieve recognition. Specifically, a linear classifier with $N$ nodes is adopted. It generates transcribed sequences of size $W/4$, where components of identical characters are ideally transcribed as repeating characters and components of non-text are transcribed as whitespace symbols. The sequence is automatically compressed into the final result. For example, $N$ is set to $37$ for English when implemented. %, meaning it is not case-sensitive.

\subsection{Vision Permutation and Architecture Variants}
There are various ways to permutate the global and Local Attention blocks in each stage. In previous researchs~\cite{du2022svtr}, it was found that placing the Local Attention block before the Global Attention block is beneficial to both feature mining of individual symbols and to guiding the Global Attention block to focus on long-term dependency capture between symbol features, while effectively saving the inference cost of the model.

As shown in Figure \ref{fig:SVIPTRarch}, in the 2-3 stages of SVIPTR, we design the structures of Local and Global Attention to process character components circularly in two ways: "series processing" and "block-parallel processing-merging", respectively, called SVIPTRv1 and SVIPTRv2. Symbolically, $[L_i][L_jG_s][G_t]$ means for each stage, Local Attention blocks are carried out at first, and then Global Attention blocks. $[L_i][L_j//G_s][G_t]$ means in the stage 2-3, Local and Global Attention blocks are carried in parallel. The $G_1$ means MHSA, the $G_2$ means ORSA, the $L_1$ means CSWin and the $L_2$ means MaSA. Others are defined similarly. Effect of the two kinds of blocks Permutations can be seen in ablation experiments.

We design several hyperparameters in SVIPTR, including channel depth and number of heads per stage, and number of mixing blocks. By varying them, SVIPTR architectures with different numbers of parameters can be obtained. In summary, we constructed four typical architectures, namely SVIPTRv1-T (Tiny), SVIPTRv2-T, SVIPTRv1-L (Large), and SVIPTRv2-B (Base). Its detailed configuration is shown in Table \ref{tab:variants}.

\begin{table}[!htbp]
        \caption{Ablation study on Attention Blocks' permutations.}
	\centering
        %\resizebox{0.95\columnwidth}{!}{
	\begin{tabular}{c|cc|cc}
		\toprule
		Atten. Permutation & IC13 & IC15 & FLOPs (G)  \\
		\midrule
            CRNN (None) &91.1 &69.4 &0.69 \\
            SVTR-T &93.5 &74.8 &0.29 \\
            \midrule
		$[L_1][L_1G_1][G_1]$ &94.4 &80 &0.269       \\  %4.03
		% $[L_1][L_1G_1][G_2]$ & 1 & 1 &0.26       \\
		$[L_1][L_1G_2][G_1]$ &94.4 &80.2 &0.258       \\ %3.99
		% $[L_1][L_1G_2][G_2]$ & 1 & 1 &0.26       \\
		
		$[L_2][L_2G_1][G_1]$ &94.3 &80.1 &0.282       \\ %4.04
		% $[L_2][L_2G_1][G_2]$ & 1 & 1 &0.26       \\
		$[L_2][L_2G_2][G_1]$ &{\bf 94.6} &{\bf 80.4} &0.27       \\ %4.01
		% $[L_2][L_2G_2][G_2]$ & 1 & 1 &0.26       \\
		
		$[L_1][L_2G_1][G_1]$ &94.1 &79.9 &0.27       \\  %4.04
		$[L_2][L_1G_1][G_1]$ &94.2 &80.1 &0.27      \\  %4.03 
		$[L_1][L_2G_2][G_1]$ &94 &80 &0.258      \\   %4.01
		$[L_2][L_1G_2][G_1]$ &94.1 &79.8 &0.258       \\ %4.0
		
		\midrule
		$[L_1][L_1//G_1][G_1]$ &92.8 &75.6 &0.195       \\ %3.24
		$[L_1][L_1//G_2][G_1]$ &92.7 &75.6 &0.192        \\ %3.23
		
		$[L_2][L_2//G_1][G_1]$ &92.9 &75.5 &0.197       \\ %3.26
		$[L_2][L_2//G_2][G_1]$ &{\bf 93.1} &{\bf 76.1} &0.194       \\ %3.24
		
		$[L_1][L_2//G_1][G_1]$ &92.3 &75 &0.196       \\ %3.24
		$[L_2][L_1//G_1][G_1]$ &92.5 &74.8 &0.196       \\ %3.25
		$[L_1][L_2//G_2][G_1]$ &92.2 &75.2 &0.193       \\ %3.24
		$[L_2][L_1//G_2][G_1]$ &92.5 &75.4 &0.193       \\ %3.24
		\bottomrule
	\end{tabular}%}
	\label{tab:permuv12}
\end{table}

\section{Experiments and Discussion}
\subsection{Experiment Settings}
\paragraph{Datasets}
Our English STR models are trained on two synthetic datasets, {\bf MJSynth (MJ)}~\cite{jaderberg2014synthetic,jaderberg2016reading} and {\bf SynthText (ST)}~\cite{gupta2016synthetic} trainsets (72k+55k), along with real-world data from {\bf COCO-Text} \cite{gomez2017icdar2017}. Testing is conducted on six public benchmarks: {\bf ICDAR 2013 (IC13)} \cite{karatzas2013icdar} testset with 857 regular images, {\bf Street View Text (SVT)}~\cite{wang2011end} testset with 647 images from Google Street View, {\bf IIIT5K-3000}~\cite{mishra2012scene} sourced from websites, {\bf ICDAR 2015 (IC15)}~\cite{karatzas2015icdar} comprising 1811 images captured via Google Glasses, {\bf Street View Text Perspective (SVTP)} \cite{phan2013recognizing} with 639 images also from Google Street View, and {\bf CUTE80 (CUTE)}, which focuses on curved text recognition, with 288 images.

For the Chinese recognition task, we employ the {\bf Chinese Scene and Document Dataset} \cite{yu2021benchmarking}.  The former includes a scene text dataset with 509,164 training, 63,645 validation, and 63,646 test images, featuring 5,880 unique characters. The document text dataset comprises 400,000 training, 50,000 validation, and 50,000 test images, with 4,865 distinct characters. We use the validation set to select the best model, which is subsequently evaluated on the test set.
% Word accuracy is used as the evaluation metric in all experiments.

\textbf{Implementations.}
The recommended training and testing configurations for SVIPTR are listed in Table \ref{tab:config}. When Training English models, data augmentation like stretch, perspective distortion, blur, and Gaussian noise, are randomly performed with probability $0.4$. Then, It was not used when training the Chinese model. All models are trained by PyTorch on four Tesla V100 GPUs.

\begin{table}[!htbp]
        \caption{Training configurations for SVIPTR.}
	\centering
	\begin{tabular}{c|c|c}
		\hline\rule{0pt}{8pt}
		Setup & English model & Chinese model  \\
		\hline\rule{0pt}{8pt}
		Rectification module  &\multicolumn{2}{c}{\cite{shi2018aster}}  \\
            \hline\rule{0pt}{8pt}
		  Distortion correction &\multicolumn{2}{c}{$32\times 64$}      \\
		\hline\rule{0pt}{8pt}
		Optimizer &\multicolumn{2}{c}{AdamW}                          \\
            \hline\rule{0pt}{8pt}
		Weight decay &\multicolumn{2}{c}{0.05}                         \\
            \hline\rule{0pt}{8pt}
		Lr scheduler &\multicolumn{2}{c}{Cosine annealing}                         \\
            \hline\rule{0pt}{8pt}
		Input size &$32\times 96$ & $32\times 320$                         \\
            \hline\rule{0pt}{8pt}
		Batch size &1024 & 512                         \\
            \hline\rule{0pt}{8pt}
		Initial learning rate &5e-4$\times \frac{batchsize}{2048}$ & 3e-4$\times \frac{batchsize}{512}$                         \\
            \hline\rule{0pt}{8pt}
		Training epochs &49 &100                         \\
            \hline\rule{0pt}{8pt}
		Warm-up epochs &2 &5                         \\
            \hline\rule{0pt}{8pt}
		Evaluation metric &\multicolumn{2}{c}{Word accuracy}                         \\
		\hline
	\end{tabular}
	\label{tab:config}
\end{table}
\textbf{Comparison with State-of-the-Art.}
As indicated in Table \ref{tab:sota}, we compare SVIPTR with prior studies on six English and one Chinese scene benchmarks. Even the fastest variant, SVIPTRv2-T, achieves highly competitive accuracy, particularly among models without Language Models (LM). In English datasets, SVIPTRv1-L outperforms LM-used models and SVTR-L, even achieving the best accuracy on IC15. Both SVIPTRv1-L and SVIPTRv2-B demonstrate overall accuracy comparable to recent studies~\cite{fang2021read,tang2021visual,lyu2022maskocr} while maintaining simplicity and faster execution.

Turning to the Chinese Scene Dataset, the accuracy of 8 existing methods is provided by~\cite{du2022svtr}. Encouragingly, SVIPTR performs notably well, with a 0.5\% accuracy gain compared to SVTR-L, the best-performing model among those listed. Other SVIPTR variants also exhibit advanced Chinese recognition accuracy. These results are attributed to SVIPTR's superior ability to perceive multi-grained character component features and its breakthroughs in characterizing Chinese words with rich stroke patterns.

\begin{table*}[ht]
        \caption{SVIPTR architecture variants (w/o counting the rectification module and CTC decoder).}
	\centering
	\begin{tabular}{c|c|c|c|c|c|c}
		\toprule
		Model & $[C_0,C_1,C_2,C_3]$ & $[N_1,N_2,N_3,N_4]$ & Heads & Attention Permutation & Params (M) & FLOPs (G)  \\
		\midrule
		SVIPTRv1-T & $[64,128,256,192]$ & $[3,3,3,3]$ & $[2,4,4,8]$ & $[L_1][L_1G_2][G_1]$ &4.0 &0.26       \\
		SVIPTRv1-L & $[192,256,512,384]$ & $[3,7,2,9]$ & $[6,8,8,16]$ & $[L_2][L_2G_2][G_1]$ &37.7 &2.31 \\
		SVIPTRv2-T & $[64,128,256,192]$ & $[3,3,3,3]$ & $[2,4,4,8]$ & $[L_1][L_1//G_2][G_1]$ &3.2 &0.19       \\
		SVIPTRv2-B & $[128,256,384, 256]$ & $[3,6,6,9]$ & $[4,8,8,12]$ & $[L_2][L_2//G_2][G_1]$ &20.2 & 1.18      \\
		\bottomrule
	\end{tabular}
	\label{tab:variants}
\end{table*}

\begin{table*}[ht]
        \caption{Results on six English and Chinese scene benchmarks tested against existing methods. The model types from top to bottom are CTC-based, attention-based, LM-based, and using dedicated decoders. * means the lightweight version of the corresponding model. The speed is the average inference time on one Tesla V100 GPU with batch size 4 (to align with~\protect\cite{du2022svtr}) for 3000 English image texts. }
	\centering
        \resizebox{0.92\textwidth}{!}{
	\begin{tabular}{l|cccccc|c|c|c|c}
		%\hline\rule{0pt}{8pt}
	   \toprule
		\multirow{2}{*}{Methods} & \multicolumn{3}{c}{English Regular} & \multicolumn{3}{c|}{English Unregular} & \multirow{2}{*}{Avg.} & \multirow{2}{*}{\makecell[c]{Chinese \\ Scene}} & \multirow{2}{*}{\makecell[c]{Params \\ (M)}} & \multirow{2}{*}{\makecell[c]{Speed $\downarrow$ \\ (ms)}} \\ 
                         & IC13       & SVT      & IIIT5k      & IC15        & SVTP        & CUTE       &                       &                                &                             &                             \\ 
            \midrule
		CRNN~\cite{shi2016end} &91.1 &81.6 &82.9 &69.4 &70.0 &65.5 &78.5 &53.4 &8.3 &6.3 \\
		Rosetta~\cite{borisyuk2018rosetta} &90.9 &84.7 &84.3 &71.2 &73.8 &69.2 &80.3 &- &44.3 &10.5 \\
		ViTSTR~\cite{atienza2021vision} &93.2 &87.7 &88.4 &78.5 &81.8 &81.3 &85.6 &- &85.5 &11.2 \\
		  SVTR-T~\cite{du2022svtr}           &96.3 &91.6 &94.4 &84.1 &85.4 &88.2 &90.7 &67.9 &6.0 &4.5     \\
		SVTR-L~\cite{du2022svtr}           &97.2 &91.7 &96.3 &86.6 &88.4 &{\bf 95.1} &92.8 &72.1 &40.8 &18.0    \\
            \midrule
		%\hline\rule{0pt}{8pt}
		% MORAN\cite{luo2019moran} &91.3 &88.3 &91.2 &68.4 &76.1 &77.4 & &83.3 &51.8 &28.5 &-         \\
		ASTER~\cite{shi2018aster} &90.8 &89.5 &93.4 &76.1 &78.5 &79.5 &86.5 &54.5 &27.2 &-   \\
		NRTR~\cite{sheng2019nrtr}  &94.7 &88.3 &86.5 &74.1 &79.4 &88.2 &84.0 &- &31.7 &160.0      \\
		SAR~\cite{li2019show}   &91.0 &84.5 &91.5 &69.2 &76.4 &83.5 &83.6 &62.5 &57.5 &120.0   \\
		SATRN~\cite{lee2020recognizing}   &95.7 &93.5 &96.1 &84.1 &88.5 &90.3 &91.9 &- &- &-       \\
		%\hline\rule{0pt}{8pt}
            \midrule
	    SRN~\cite{yu2020towards}   &95.5 &91.5 &94.8 &82.7 &85.1 &87.8 &90.4 &60.1 &54.7 &25.4         \\
		
		% ABINet*\cite{fang2021read} &94.9 &90.4 &94.6 &81.7 &84.2 &86.5 &&89.8 &- &23.5 &50.6       \\
		ABINet~\cite{fang2021read} &97.4 &93.5 &96.2 &86.0 &89.3 &89.2 &92.6 &- &36.7 &51.3       \\
		VisionLAN~\cite{wang2021two} &95.7 &91.7 &95.8 &83.7 &86.0 &88.5 &91.2 &- &32.8 &28.0          \\
		% VST~\cite{tang2021visual}     &96.4 &93.8 &96.3 &85.4 &88.7 &{\bf 95.1} &92.6 &- &64.0 &-         \\
		MATRN~\cite{na2022multi}     &95.8 &{\bf 94.9} &{\bf 96.7} &82.9 & 90.5 &94.1 &92.3 &- &44.2 &29.6         \\
            MaskOCR*~\cite{lyu2022maskocr}   &97.7 &93.7 &95.4 &86.6 &89.0 &87.5 &92.5 &68.6 &36.0 &24.1        \\
            MaskOCR~\cite{lyu2022maskocr}   &96.8 &94.7 &95.3 &87.1 &89.3 &90.6 &92.7 &68.8 &100.0 &-        \\
            % TrOCR*\cite{li2023trocr} &97.3 &91 &90.1 &81.1 &90.7 &86.8 &&88.7 &- &334 &19.8 \\
            TrOCR~\cite{li2023trocr} &{\bf 98.3} &93.2 &91.0 &84.0 &{\bf 91.0} &89.6 &90.2 &- &558.0 &20.8 \\
		%\hline\rule{0pt}{8pt}
            \midrule
		MGP-STR-T~\cite{wang2022multi} &96.5 &93.2 &96.3 &86.2 &89.4 &90.6 &92.7 &- &5.4 &10.6      \\
            MGP-STR(Fuse)~\cite{wang2022multi} &97.3 &94.7 &96.4 &87.2 &91.0 &90.2 &93.3 &- &21.0 &12.0  \\
            SVTR-L+DCTC~\cite{zhang2023self} &97.4 &93.7 &96.9 &87.3 &88.5 &92.3 &93.3 &73.9 &40.8 &18.0  \\
		% PREN*~\cite{yan2021primitive} &94.7 &92.0 &92.1 &79.2 &83.9 &81.3 &88.0 &- &29.1 &40.0     \\
		% PREN2D\cite{yan2021primitive} &96.4 &94 &95.6 &83 &87.6 &91.7 &&91.5 &- &- &-         \\
		
		%\hline\rule{0pt}{8pt}
            \midrule
		SVIPTRv1-T (Tiny) &96.8 &93.5 &93.3 &87.2 &87.1 &87.4 &91.4 &68.7 &5.9 &4.2         \\
		SVIPTRv2-T       &95.8 &92.6 &92.6 &85.4 &86.0 &85.0 &90.3 &65.8 &\bf{5.1} &\bf{3.3}         \\
		SVIPTRv1-L (Large) &97.3 &94.5 &95.1 &{\bf 88.4} &89.6 &88.8 &{\bf 92.9} &{\bf 72.6} &39.5 &16.9         \\
		SVIPTRv2-B (Base)  &97.0 &94.2 &94.8 &88.0 &90.0 &90.2 &92.7 &72.0 &22.1 & 8.0        \\
            \midrule
            SVIPTRv1-L+BiLSTM+DCTC &98.0 &94.8 &96.1 &{\bf 88.8} &90.4 &91.2 &{\bf 93.6} &{\bf 74.2} &45.8 &22.0\\
            %\hline
		\bottomrule
	\end{tabular}}
	\label{tab:sota}
\end{table*}
% \subsection{Further Analysis}
\subsection{Ablation Study}
%\paragraph{Ablation Experiments}
To gain a deeper insight into SVIPTR, we conducted controlled experiments on both IC13 (regular) and IC15 (irregular) under various configurations. The training set and experimental Settings in the English scenario are unchanged. In addition, we conduct a series of comparative experiments to verify the effectiveness of our models on image texts of varying lengths. The Chinese Document Dataset is used for training and validation. Three test sets are employed: Testset1, the standard set from the Chinese Document Dataset; Testset2, consisting of 2,367 images cropped from real documents with lengths between 320 and 640 pixels; and Testset3, featuring 638 images, also derived from real documents but with lengths exceeding 640 pixels. For efficiency, all the experiments are carried out by using SVIPTR-T without the rectification module and data augmentation.
% The English dataset used in the experiment consists of COCO-Text and part of the real scene training set, with a total of 45054 images over 500 epochs of training.
% The training set and experimental Settings in the English scenario are unchanged.

\textbf{Effective Permutation of Attention Blocks.}
The top half and the bottom half of Table \ref{tab:permuv12} respectively show the performance of SVIPTRv1-T and SVIPTRv2-T composed of various Local and Global Attention and their permutations.
% $[L_i][L_jG_s][G_t]$ means for each stage, Local Attention blocks are carried out at first, and then Global Attention blocks. $[L_i][L_j//G_s][G_t]$ means in the stage 2-3, Local and Global Attention blocks are carried in parallel. The $G_1$ means MHSA, the $G_2$ means ORSA, the $L_1$ means CSWin and the $L_2$ means MaSA. Others are defined similarly. 
It can be observed that almost every scheme gives a certain accuracy improvement compared to CRNN. We believe that the improvements are attributed to the comprehensive perception of character components features by the self-attention mechanism. The relatively large gains on irregular text further explain the self-attention mixing block is helpful to feature learning in complex scenarios. It is observed that $[L_2][L_2G_2][G_1]$ reports the best accuracy in the top half of Table \ref{tab:permuv12} and $[L_2][L_2//G_2][G_1]$ achieves best in the bottom half of Table \ref{tab:permuv12}. The best one gives accuracy gains of 1.4\% and 2.8\% on IC13 and IC15 when compared with SVTR-T. It is observed that $[L_1][L_1//G_2][G_1]$ reports the fastest inferencing speed. In general, SVIPTR has better performance than SVTR without using rectification module, and it is the fastest and highest performance visual feature encoder.

%\subsubsection{Effectiveness of Positional Embeddings.}
\textbf{Effectiveness of Positional Embeddings.}
As indicated in Table \ref{tab:peffn}, various PE schemes exhibit slight variations in recognition accuracy. The LePE scheme surpasses the default one (None) by 2.4\% and 5.3\% on the two datasets, demonstrating its effectiveness, particularly in the case of irregular text. The use of LePE not only improves the recognition accuracy without affecting the inference efficiency but also is insensitive to the input length, which enhances the versatility of SVIPTR.

\begin{table}[!htbp]
        \caption{Ablation study on Positional Embeddings.}
        \setlength{\tabcolsep}{3pt} % 设置列间距为10pt
	\centering
	\begin{tabular}{ccc|cccc}
		\toprule
		Model + Positional Embedding & IC13 & IC15 & FLOPs (G) \\
		\midrule
		SVIPTRv1-T + None & 93.7 & 79.7  & 0.25    \\
		SVIPTRv1-T + APE & 94 & 78.2 & 0.26       \\
		
		SVIPTRv1-T + CPE &94.3 &78.9 & 0.26       \\
		SVIPTRv1-T + LePE &{\bf 94.6} &{\bf 80.3}  & 0.26       \\
		
		\bottomrule
	\end{tabular}
	\label{tab:peffn}
\end{table}
% \subsubsection{Results for different lengths inputs testing}
\textbf{Results for different lengths inputs testing.}
In Table \ref{tab:dinput}, the performance of the three models on the three test sets is quite different. SVTR's inference performance starts to collapse as the input length gets larger. SVTR uses APE and SWin self-attention with static Windows, which does not have CNN-like translation invariance. So the length of inference images are limited by the size of training images. If inference images' lengths are beyond the pre-set, resizing the image is required, which results in distortion. SVIPTR is not only length-insensitive, but the inference performance will not decrease as the input length increases.
% SVIPTR not only supports arbitrary lengths of image text input, but the inference performance will not decrease as the input length increases.
\begin{table}[!htbp]
        \caption{Ablation studies on inputs of different lengths.}
	\centering
	\begin{tabular}{c|c|c|c}
		\toprule
		Model & Testset1 & Testset2 & Testset3 \\
		\midrule
		CRNN &97.5 &91 &91.6       \\
		SVTR-T &98.2 & 47.5 &5.9       \\
		
		SVIPTRv1-T &{\bf 98.8} &{\bf 92.1} &{\bf 93.7}       \\
		
		\bottomrule
	\end{tabular}
	\label{tab:dinput}
\end{table}

\subsection{Attention Visualization}
%我们已经将SVIPTR在解码不同字符组件时的注意图可视化。最终选择用所提出的性能最弱的SVIPTRv2-T模型来生成注意图并在本文中展示，进而佐证SVIPTR所有变体的高性能。如图4所示，第一行的图片是原图，右侧为模型预测结果。下面的注意图可以解释为在整个识别过程中发挥作用的全局和局部注意力机制，其中第一张注意图表示在识别过程中对字符"e"的注意程度，从这里可以看出模型可以在比较长的时间跨度内对目标字符保持很强的关注，进而验证了全局注意力机制的有效性。后续的注意图是为原图中每个字符组件选取一个合适的时间戳对应的注意力而组成的，从几乎每个字符都被亮色覆盖这一现象，可以看出模型在每个字符组件的局部附近都分配了足够的注意力，验证了局部注意力机制的有效性，并且注意力不会随着字符序列变长而明显衰减，这意味着成功捕获了不同字符组件之间的依赖关系，进而说明模型有良好的长距离推理能力。图4表明，SVIPTRv2-T捕获了字符局部、整个字符和跨字符的多种特征和线索，这与SVIPTR 感知多粒度字符组件特征的说法一致。这再次说明了SVIPTR及其变体的有效性。

\begin{figure}[h]
	\centering
	\includegraphics[width=1.0\linewidth]{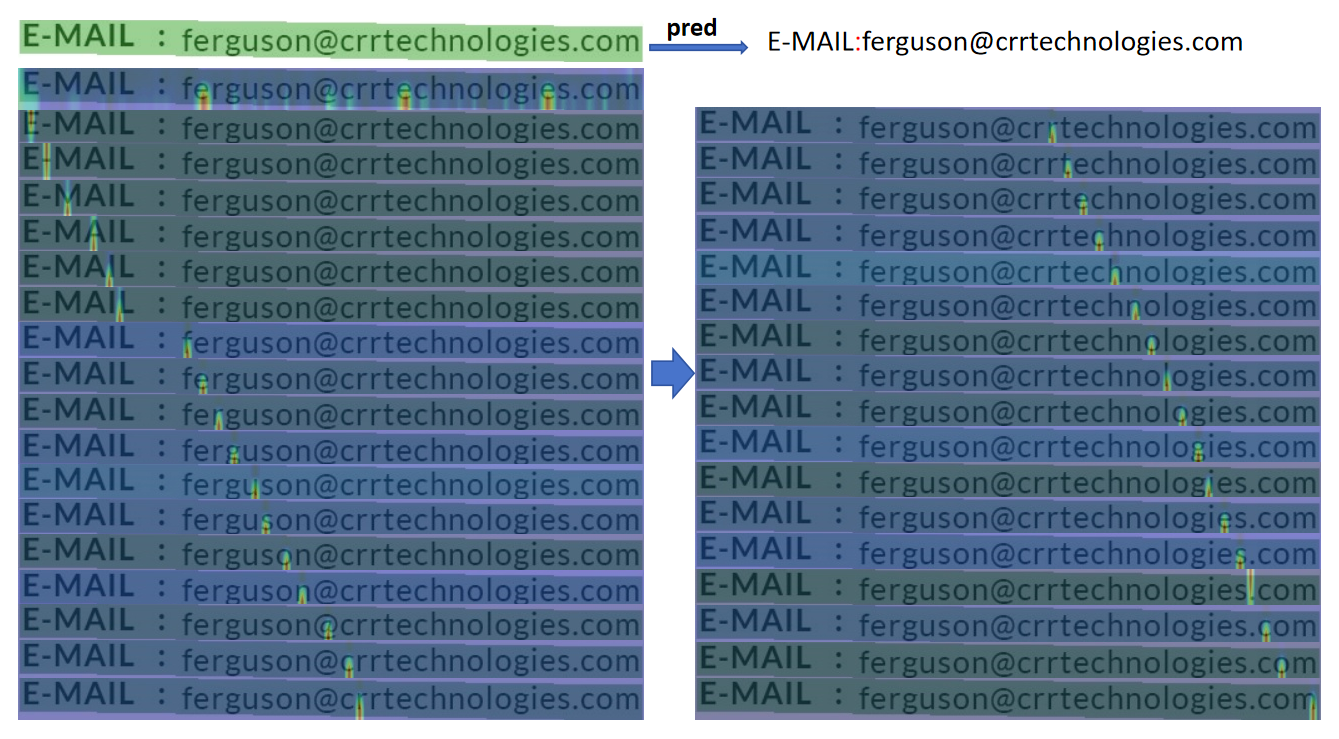}
	\caption{Visualization of the attention maps for SVIPTRv2-T. The character in red means it is missed or an error.}
	\label{fig:longattn}
        \Description{Visualization of the attention maps for SVIPTRv2-T. The character in red means it is missed or an error.}
\end{figure}

We have visualized the attention maps of SVIPTR when decoding different character components. Finally, the proposed SVIPTRv2-T model without TPS and STN (It is obtained by training a large number of image texts with a uniform shape size of 32x96) with the weakest performance is selected to generate the attention map and shown in this paper, which further proves the high performance of all variants of SVIPTR. As shown in Figure \ref{fig:longattn}, the picture in the first row is the original image (the shape size is $32\times 480$), and the right side is the model prediction result. The following attention map can be interpreted as the global and local attention mechanisms that play a role in the whole recognition process. The first attention map represents the degree of attention given to the character "e" in the recognition process. From here, it can be seen that the model can maintain strong attention to the target character in a relatively long time span, which verifies the effectiveness of the global attention mechanism. The subsequent attention map is composed of selecting an appropriate timestamp corresponding to the attention of each character component in the original image. From the phenomenon that almost each character is covered by a bright color, it can be seen that the model allocates enough attention to the local vicinity of each character component, which verifies the effectiveness of the local attention mechanism. Moreover, the attention does not significantly decay as the character sequence becomes longer, which means that the dependencies between different character components are successfully captured, which further indicates that the model has good long-distance reasoning ability. 

Figure \ref{fig:longattn} indicates that SVIPTRv2-T captures multiple features and cues locally, throughout, and across characters, which is consistent with the claim that SVIPTR senses multi-granularity character component features. It also verifies that the inference of SVIPTR is not sensitive to the length of the image, and the weakest performing variant also has strong long-distance inference ability. This again illustrates the effectiveness of SVIPTR and its variants.

\section{Conclusion}
In this study, we propose SVIPTR, an innovative vision permutable extractor for fast and efficient Scene Text Recognition (STR). SVIPTR excels at extracting and fusing visual and semantic features of image text, and it is insensitive to the length of the inference input. These grants SVIPTR notable accuracy, efficiency, and cross-lingual adaptability. We have also developed different versions of SVIPTR to cater to a range of application requirements.  Experiments on both English and Chinese benchmarks generally validate the effectiveness of our models. It achieves highly competitive or superior accuracy compared to state-of-the-art methods while maintaining faster processing speeds. 
Our future goal is to integrate SVIPTR with an efficient text decoder to further elevate STR performance. 
% SVIPTR excels at extracting character features and capturing contextual dependencies between character components, all while being unrestricted by input size during inference.

% \begin{acks}
% \end{acks}

% \section*{Acknowledege}
\begin{acks}
This work was supported in part by the National Natural Science Foundation of China (Grant Nos. U1636211, U2333205, 61672081, 62302025, 62276017), a fund project: State Grid Co., Ltd. Technology R\&D Project (ProjectName: Research on Key Technologies of Data Scenario-based Security Governance and Emergency Blocking in Power Monitoring System, Proiect No.: 5108-202303439A-3-2-ZN), the 2022 CCF-NSFOCUS Kun-Peng Scientific Research Fund and the Opening Project of Shanghai Trusted Industrial Control Platform and the State Key Laboratory of Complex \& Critical Software Environment (Grant No. SKLSDE-2021ZX-18).
\end{acks}

\bibliographystyle{ACM-Reference-Format}
\bibliography{sample-base}

\appendix

\end{document}